\documentclass[10pt,twocolumn,letterpaper]{article}

\usepackage{cvpr}
\usepackage{times}
\usepackage{epsfig}
\usepackage{graphicx}
\usepackage{amsmath}
\usepackage{enumitem}
\usepackage{amssymb}
\usepackage{xcolor}
\usepackage{booktabs}
\usepackage{algorithmic}
\usepackage{verbatim}

\usepackage[breaklinks=true,bookmarks=false]{hyperref}

\cvprfinalcopy 

\def\cvprPaperID{****} 

\begin{document}

\title{Multi-Task Multicriteria Hyperparameter Optimization}

\author{Kirill Akhmetzyanov\\
Perm State Technical University\\
{\tt\small KRAkhmetzjanov@pstu.ru}
\and
Alexander Yuzhakov\\
Perm State Technical University\\
{\tt\small uz@at.pstu.ru}
}

\maketitle

\begin{abstract}
   We present a new method for searching optimal hyperparameters among several tasks and several criteria. Multi-Task Multi Criteria method (MTMC) provides several Pareto-optimal solutions, among which one solution is selected with given criteria significance coefficients. The article begins with a mathematical formulation of the problem of choosing optimal hyperparameters. Then, the steps of the MTMC method that solves this problem are described. The proposed method is evaluated on the image classification problem using a convolutional neural network. The article presents optimal hyperparameters for various criteria significance coefficients.
\end{abstract}

\section{Introduction}

Hyperparametric optimization~\cite{Hutter2009} is an important component in the implementation of  machine learning models (for example, logistic regression, neural networks, SVM, gradient boosting, etc.) in solving various tasks, such as classification, regression, ranking, etc. The problem is how to choose the optimal parameters when a trained model is evaluated using several sets and several criteria.

This article describes an approach to solving the problem described above. We will present the results of experiments on the selection of hyperparameters obtained using the proposed approach (MTMC) with various criteria significance coefficients.

The article is organized as follows. First, we discuss related work in Section 2. Section 3 describes the proposed method. Section 4 presents the results of experiments on the selection of optimal hyperparameters. Section 5 contains the conclusion and future work.

\section{Related Work}

The problem of choosing optimal hyperparameters has long been known. Existing methods for solving this problem give both the single optimal solution, and several ones. 

In~\cite{Sener2018}, a Pareto optimization method is proposed, in which the optimal solution is given for several problems simultaneously. This method consists in minimizing the weighted sum of loss functions for each of the tasks. \cite{Fliege2000} describes the Pareto optimization method, which gives an optimal solution according to several criteria based on gradient descent, and this optimization is also carried out in the learning process. In~\cite{Igel2005}, the search for a Pareto-optimal solution is carried out according to several criteria. The method in~\cite{Bengio2000} gives optimal hyperparameters using back propagation through the Cholesky decomposition. In~\cite{Bergstra2011}, optimization is performed using a random choice of hyperparameters based on the expected improvement criterion. \cite{Bergstra2012} proposes method of hyperparametric optimization based on random search in the space of hyperparameters. In~\cite{Snoek2012}, search for optimal hyperparameters is carried out using Bayesian optimization.

The novelty of MTMC method:

1. Optimization is carried out simultaneously according to several criteria and several tasks with setting the significance of the criteria.

2. The choice of optimal hyperparameters is provided \textit{after} training and evaluation, which eliminates the need to re-train the model.

3. The proposed method does not need to be trained.

\section{Optimization of Hyperparameters According to Several Tasks and Several Criteria}

First, we describe the mathematical problem that MTMC solves, then we present the steps performed in MTMC.

\subsection{Formalization of the problem}

In the proposed method, the model is evaluated on several test sets (tasks) $  T $. The problem of finding a minimum for tasks $ T $ is known as minimizing the expected value of empirical risk~\cite{Vapnik1992}.

The choosing optimal hyperparameters is formalized as follows:
\begin{equation}
\theta=\underset{\theta \in \Theta}{\operatorname{argmin}} \mathbb{E}_{\tau}[\mathcal{L}(\theta, \varphi)]
\end{equation}
where $ \Theta $ – the set of all hyperparameters, $ \theta $  – the selected optimal hyperparameters, $ \varphi $ – the vector of significance coefficients of the criteria, $ \mathcal{L(\cdot)} $ – the estimation function of the model with the given hyperparameters $ \theta $ and the coefficients $ \varphi $, $ \tau $ – the task for which optimization is performed.

The developed method gives a solution to the problem~(1).

\subsection{Description of MTMC}

According to~(1), the developed method should fulfill the following requirements:

1) the method should solve the minimization problem;

2) the significance of each criterion is determined by the vector of coefficients $ \varphi $ (the higher the coefficient, the more important the corresponding criterion).

We denote the test sample of the task $ \tau $:
\begin{equation}
x^{i} \sim \mathcal{D}, i=1 \ldots N_{\text {task }}
\end{equation}
where $ x^{i} $ – the $ i^{th} $ test set has the distribution $ \mathcal{D} $, $ N_{\text {task }} $ – the number of tasks.
	
Before choosing hyperparameters, for model $ \mathcal{M} $ we obtain an evaluation matrix for the test set  $ x^{i} $ and the given evaluation criteria:
\begin{equation}
V=\mathcal{M}\left(x^{i} ; \Theta\right)
\end{equation}
\begin{equation}
\mathcal{M}\left(x^{i} ; \Theta\right)::\left(\mathbb{R}^{x_{\text {sise}}}, \mathbb{R}^{N_{\text {parameter}} \times N_{\text {combination}}}\right) \rightarrow \mathbb{R}^{N_{\text {combination}} \times N_{\text {criteria}}}
\end{equation}
where $ \mathcal{M}(\cdot) $ – the model function that transforms the given set $ x^{i} $ and with the given hyperparameters $ \Theta $ into the evaluation matrix $ {V} $, $ N_{\text {criteria}} $ – the number of criteria, $ x_{\text {size}} $ –  the dimension of the test set, $ N_{\text {parameter}} $ – the number of hyperparameters, $ N_{\text {combination}} $ –  the number of hyperparameter combinations.

Then, the function $ \mathcal{L} $ is calculated for each set $ x^{i} $, which is formally described as follows:
\begin{equation}
\mathcal{L}(\cdot ; \mathbf{\Theta}, \varphi)=\mathcal{E}(V ; \varphi)
\end{equation}
\begin{equation}
\mathcal{E}(V ; \varphi)::\left(\mathbb{R}^{N_{\text {criteria }}}, \mathbb{R}^{N_{\text {criteria }}}\right) \rightarrow \mathbb{R}^{1}
\end{equation}

MTMC method gives Pareto optimal solutions in which the following steps are performed:

1. The vectors from the evaluation $ {V} $ (the number of such vectors is $ N_{\text {criteria}} $) is in the space of given criteria.

2. Then we get Pareto optimal solutions $\tilde{V} \subseteq V$ – the nearest Pareto front to the origin of the criteria space:
\begin{equation}
\tilde{V}=\text { ParetoFront }(V), \quad \tilde{V} \in \mathbb{R}^{N_{\text {opt }} \times N_{\text {criteria}}}
\end{equation}
where $ N_{\text {opt}} $ – the number of Pareto optimal solutions.

3. The optimal solutions $\tilde{v} \in \tilde{V}$ are scaled according to each criterion to the interval $[0 ; 1]$:
\begin{multline}
\tilde{V}_{\text {scaled}}=\frac{\tilde{V}_{i}-\tilde{v}_{\text {min}}}{\tilde{v}_{\text {max}}-\tilde{v}_{\text {min}}}, \tilde{v}_{\text {min}} \in \mathbb{R}^{N_{\text {criteria}}}, \\ \tilde{v}_{\text {min}} \in \mathbb{R}^{N_{\text {criteria}}}, i=1 \ldots N_{opt}
\end{multline}
where $\tilde{v}_{max}$ – the vector of maximum values of $\tilde{V}$ for each criterion, $\tilde{v}_{min}$ – the vector of minimum values of $\tilde{V}$ for each criterion.

Thus, the optimal solution is the solution closest to the origin, and if any solution $\tilde{v} \in \tilde{V}$ is the origin, then it is optimal for any $ \varphi $.

4. The vector $ \varphi $ in the space of criteria is defined.

We introduce the vector of the optimal solution, which is the middle of the segment $ [0 ; 1] $ in the axes of the criteria space:
\begin{equation}
\varphi_{opt}=\left(\forall i: \varphi_{0}=\cdots=\varphi_{i}=\cdots=\varphi_{N_{ {criteria}}}=\frac{1}{2}\right)
\end{equation}

Conditions for $ \varphi $ are:
\begin{equation}
\varphi=\left\{\begin{array}{c}
{\varphi_{\text {opt }}, \text { if } \forall i: \varphi_{i}=0}, \\
{\varphi \in[0 ; 1], \text { otherwise. }}
\end{array}\right.
\end{equation}

5. Project the vectors from the matrix $ \tilde{V}_{scaled} $ onto the vector $ \varphi $:
\begin{equation}
\tilde{V}_{proj}=\frac{{\tilde{V}_{scaled}}^{T} \cdot \varphi}{\|\varphi\|}, \quad \tilde{V} \in \mathbb{R}^{N_{o p t}}.
\end{equation}

From~(9) and~(11) it follows that if the vectors $ \varphi $ and $ \varphi_{opt} $ are collinear, then:
\begin{multline}
 \exists \lambda: \varphi=\lambda \cdot \varphi_{opt} \Rightarrow \tilde{V}_{proj}=\sum_{i}\left[\frac{1}{\varphi_{i}} \cdot \frac{\tilde{V}_{scaled_{i}}}{\|\varphi\|}\right] \propto \\ \propto \sum_{i} \tilde{V}_{scaled_{i}}=\left\|\tilde{V}_{scaled}\right\|_{1}. 
\end{multline}

That is, in the case of equality of all elements of $ \varphi $, the minimization problem reduces to finding the minimum $ {L1} $-norm $ \tilde{V}_{scaled} $.

From~(11) it also follows that if some component of the vector $ \varphi $ is equal to zero, then the corresponding criterion will not affect the choice of the optimal hyperparameter. If all criteria are equal to zero, except for one, then only the criterion with a nonzero component of the vector $ \varphi $ will affect the choice of optimal hyperparameters.

6. We find hyperparameters $ \theta $ at which the minimum of the vector $ \tilde{V}_{proj} $ is reached:
\begin{equation}
\theta=\underset{\theta}{\operatorname{argmin}} \tilde{V}_{\text {proj}}
\end{equation}

\begin{figure}[t]
	\begin{center}
		\includegraphics[width=0.8\linewidth]{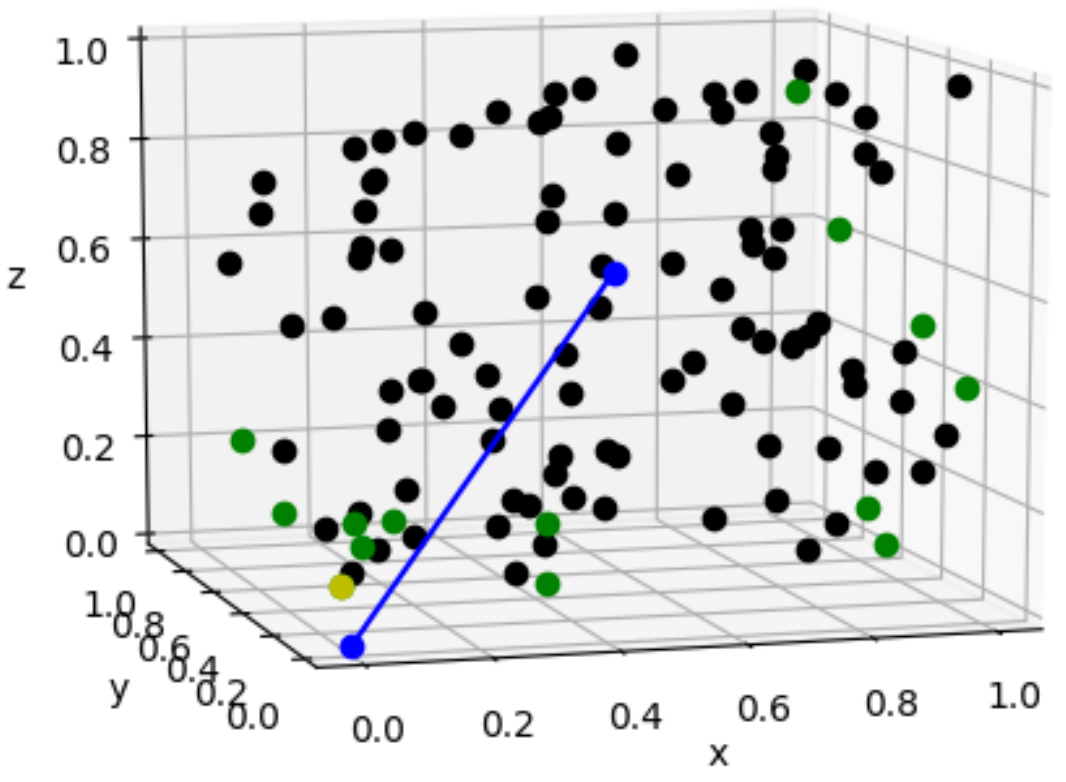}
	\end{center}
	\caption{Example of a solution given by MTMC, green points denote Pareto optimal solutions, blue vector is the vector $ \varphi $, yellow point denotes the optimal solution given by MTMC for a given $ \varphi $.}
	\label{fig:mtmcexample}
\end{figure}

Fig.~\ref{fig:mtmcexample} shows an example solution using MTMC for random numbers in the three-dimensional space.

\section{Conducting experiments}

First, the evaluation matrix $ {V} $ for the selected model $ \mathcal{M} $ is obtained. Then, for various combinations of components of $ \varphi $, optimal hyperparameters are selected using MTMC.

\subsection{Obtaining the evaluation matrix}

The developed MTMC method is applied to solve the problem of image classification. The problem we are solving is described in the article~\cite{Akhmetzyanov2019}.

In~\cite{Akhmetzyanov2018}, we selected  the MobileNet neural network architecture~\cite{Howard2017} as a mathematical model for image processing.

The search for optimal hyperparameters was carried out among two popular training methods: changing the learning speed based on the epoch $ lr=base\_lr \cdot lr\_decay^{evoch} $ (where $ base\_lr $ is the initial learning rate, $ lr\_decay $ is the coefficient of decreasing learning rate, epoch is the number of epochs) and cyclical learning~\cite{Smith2017}. In cyclic learning, there are three ways to change the learning rate:

1. triangular – fixed initial learning rate (\textit{base\_lr}), maximum fixed learning rate (\textit{max\_lr}), learning rate increases from \textit{base\_lr} to \textit{max\_lr} and decreases from \textit{max\_lr} to \textit{base\_lr} linearly.

2. triangular2 – fixed initial learning rate (\textit{base\_lr}), maximum learning rate (\textit{max\_lr}), learning rate, as in triangular, varies linearly, but \textit{max\_lr} in the learning process is halved.

3. exp\_range – fixed initial learning rate (\textit{base\_lr}), maximum learning rate (\textit{max\_lr}), learning rate also changes linearly, but \textit{max\_lr} in the learning process decreases exponentially.

In the first learning method, the hyperparameters are the value of the initial learning rate (\textit{base\_lr}) and the coefficient of decreasing learning rate (\textit{lr\_decay}). In the second method, hyperparameters – a way to change the learning rate (\textit{cyclic\_mode}), the value of the initial learning rate (\textit{base\_lr}) and maximum learning rate (\textit{max\_lr}).

For each hyperparameter, a range of change and a constant step of change within the range were selected. For training, Grid search was used among $ N_{combination}=100 $ combinations of hyperparameters.

For each combination of hyperparameters, training was carried out using cross-validation k-fold~\cite{Stone1974} with 10 folds. For training, Keras framework was used~\cite{keras}. The training lasted 15 epochs; the test was carried out on $ N_{task}=5 $ different test sets. That is, $ 100 \cdot 10 = 1000 $ is number of different neural networks, $ 15 \cdot 1000 = 15000 $ neural networks evaluations are conducted, $ 15000 \cdot 5 = 75000 $ evaluation results are obtained. Neural networks trained on ten TPUs~\cite{tpu}, which took several days.

Among all epochs, for each fold and for each test set, the maximum accuracy is selected, as well as the number of the epoch at which the maximum accuracy is achieved. The following values are calculated for each test set among the folds: the expected value and the variance of the classification error, the expected value and the variance of the epoch number at which convergence on the test set is achieved. We have obtained an evaluation matrix among all neural networks with their hyperparameters and among all test samples.

\subsection{Processing the evaluation matrix}

Based on~(1), for each criterion, among all the samples, the expected value is considered. That is, for all test sets, the criteria: (i)~the sample mean of the classification error, (ii)~the sample variance of the classification error, (iii)~the sample mean and (iv)~sample variance of the epoch number at which convergence is achieved in the test sample. These values are the criteria for evaluating hyperparameters for a certain test set (matrix $ {V} $ from~(3)) with the number of criteria $ N_{criteria}=4 $.

$ \tilde{V} $ is calculated from~(7), the number of Pareto optimal solutions obtained is $ N_{opt}=25 $. Optimal hyperparameters, i.e., $ \tilde{V} $, are presented in~Appendix~A.

The vector of the optimal solution according to~(9) for $ N_{criteria}=4 $ is $ \varphi_{opt}=\{0.5 ; 0.5 ; 0.5 ; 0.5\} $. Next, calculations are carried out according to~(8) and~(11), and for various $ \varphi $ optimal solutions are chosen according to~(5). These optimal solutions are presented in~Appendix~B. Also,~Appendix~C shows the graphs of changes in the accuracy of the neural network for each test set and each epoch for the obtained optimal hyperparameters.

\section{Conclusion}
In this work, we proposed a new method for hyperparameter optimization among several tasks and several criteria. We trained several neural networks with various hyperparameters to solve the image classification problem. Then, for these neural networks, evaluation matrices were obtained on several tasks. We applied MTMC to these matrices and got optimal solutions with different significance coefficients. In the future, we will work to create a meta-learning method that solves the same problem as the method described in this article, but optimization will be performed among various models.
\\

\textbf{Acknowledgments:} The reported study was partially supported by the Government of Perm Krai, research project No. C-26/174.6.

{\small
\bibliographystyle{ieee_fullname}
\bibliography{egbib}
}

\appendix
\section{Pareto Optimal Hyperparameters }
Table~\ref{tab:pareto_1} and Table~\ref{tab:pareto_2} show the Pareto optimal hyperparameters for the two learning methods.
\begin{table}[!h]
	\centering
	\begin{tabular*}{\columnwidth}{p{0.5\columnwidth}|p{0.5\columnwidth}}
		\toprule[0.2em]
		base\_lr & lr\_decay \\ \toprule[0.2em]
		0.001    & 0.75      \\ \hline
		0.001    & 0.8       \\ \hline
		0.005    & 0.75      \\ \hline
		0.01     & 0.9       \\ \hline
		0.01     & 0.95      \\ \hline
	\end{tabular*}
	\vspace{1pt}
	\caption{Pareto optimal solutions for the first learning method. }
	\label{tab:pareto_1}
\end{table}
\begin{table}[!h]
	\centering
	\vspace{0pt}
	\begin{tabular*}{\columnwidth}{p{0.33\columnwidth}|p{0.33\columnwidth}|p{0.33\columnwidth}}
		\toprule[0.2em]
		base\_lr & max\_lr & cyclic\_mode \\ \toprule[0.2em]
		0.0001   & 0.005   & exp\_range   \\ \hline
		0.0001   & 0.005   & triangular2  \\ \hline
		0.0005   & 0.001   & exp\_range   \\ \hline
		0.0005   & 0.005   & triangular2  \\ \hline
		0.001    & 0.0001  & triangular2  \\ \hline
		0.001    & 0.0005  & triangular   \\ \hline
		0.001    & 0.001   & exp\_range   \\ \hline
		0.001    & 0.005   & triangular2  \\ \hline
		0.005    & 0.0001  & triangular   \\ \hline
		0.005    & 0.005   & triangular   \\ \hline
		0.01     & 0.0001  & triangular   \\ \hline
		0.01     & 0.0001  & triangular2  \\ \hline
		0.01     & 0.005   & triangular   \\ \hline
		0.01     & 0.005   & triangular2  \\ \hline
		0.01     & 0.01    & triangular   \\ \hline
		0.0001   & 0.0001  & triangular   \\ \hline
		0.0005   & 0.001   & triangular   \\ \hline
		0.0005   & 0.01    & triangular2  \\ \hline
		0.001    & 0.005   & triangular   \\ \hline
		0.005    & 0.01    & triangular   \\ \hline
	\end{tabular*}
	\vspace{1pt}
	\caption{Pareto optimal solutions for the first learning method. }
	\label{tab:pareto_2}
\end{table}
\newpage
\section{MTMC Optimal Hyperparameters}
Table~\ref{tab:mtmc_opt} shows the optimal hyperparameters $ \theta $ obtained using MTMC method for given significance coefficients of the criteria $ \varphi $.
\begin{table}[t]
	\centering
	\vspace{0pt}
	\begin{tabular*}{\columnwidth}{p{0.07\columnwidth}|p{0.07\columnwidth}|p{0.07\columnwidth}|p{0.07\columnwidth}|p{0.5\columnwidth}}
		\toprule[0.2em]
		$ \varphi_0 $ & $ \varphi_1 $ & $ \varphi_2 $ & $ \varphi_3 $ & $ \theta $                                               \\ \toprule[0.2em]
		0.5        & 0.5        & 0.5        & 0.5        & base\_lr=0.01,\newline max\_lr=0.01,\newline cyclic\_mode=triangular     \\ \hline
		0.0        & 0.5        & 0.5        & 0.5        & base\_lr=0.01,\newline max\_lr=0.01,\newline cyclic\_mode=triangular     \\ \hline
		1.0        & 0.5        & 0.5        & 0.5        & base\_lr=0.01,\newline max\_lr=0.01,\newline cyclic\_mode=triangular     \\ \hline
		0.5        & 0.0        & 0.5        & 0.5        & base\_lr=0.005,\newline max\_lr=0.0001,\newline cyclic\_mode=triangular  \\ \hline
		0.5        & 1.0        & 0.5        & 0.5        & base\_lr=0.01,\newline max\_lr=0.01,\newline cyclic\_mode=triangular     \\ \hline
		0.5        & 0.5        & 0.0        & 0.5        & base\_lr=0.01,\newline max\_lr=0.01,\newline cyclic\_mode=triangular     \\ \hline
		0.5        & 0.5        & 1.0        & 0.5        & base\_lr=0.01,\newline max\_lr=0.01,\newline cyclic\_mode=triangular     \\ \hline
		0.5        & 0.5        & 0.5        & 0.0        & base\_lr=0.0001,\newline max\_lr=0.005,\newline cyclic\_mode=triangular2 \\ \hline
		0.5        & 0.5        & 0.5        & 1.0        & base\_lr=0.01,\newline max\_lr=0.01,\newline cyclic\_mode=triangular     \\ \hline
		0.0        & 0.0        & 0.5        & 0.5        & max\_lr=0.005,\newline lr\_decay=0.75                            \\ \hline
		1.0        & 1.0        & 0.5        & 0.5        & base\_lr=0.01,\newline max\_lr=0.01,\newline cyclic\_mode=triangular     \\ \hline
		0.5        & 0.5        & 0.0        & 0.0        & base\_lr=0.0001,\newline max\_lr=0.005,\newline cyclic\_mode=triangular2 \\ \hline
		0.5        & 0.5        & 1.0        & 1.0        & base\_lr=0.01,\newline max\_lr=0.01,\newline cyclic\_mode=triangular     \\ \hline
		1.0        & 0.0        & 0.0        & 0.0        & base\_lr=0.0001,\newline max\_lr=0.005,\newline cyclic\_mode=triangular2 \\ \hline
		0.0        & 1.0        & 0.0        & 0.0        & base\_lr=0.0001,\newline max\_lr=0.005,\newline cyclic\_mode=triangular2 \\ \hline
		0.0        & 0.0        & 1.0        & 0.0        & max\_lr=0.005,\newline lr\_decay=0.75                            \\ \hline
		0.0        & 0.0        & 0.0        & 1.0        & base\_lr=0.0005,\newline max\_lr=0.001,\newline cyclic\_mode=exp\_range  \\ \hline
	\end{tabular*}
	\vspace{1pt}
	\caption{Optimal hyperparameters for relevant criteria significance coefficients. }
	\label{tab:mtmc_opt}
\end{table}

\section{Accuracy of MTMC Optimal Solutions}
The figures below show optimal solutions chosen by MTMC method: 95\% confidence intervals of the dependence of accuracy on the fold / epoch and maximum accuracy for each of the folds (\textbf{left}) and “box plot” with maximum accuracy for all test sets (\textbf{right}).
\begin{figure*}[t]
\centering
\includegraphics[width=\linewidth]{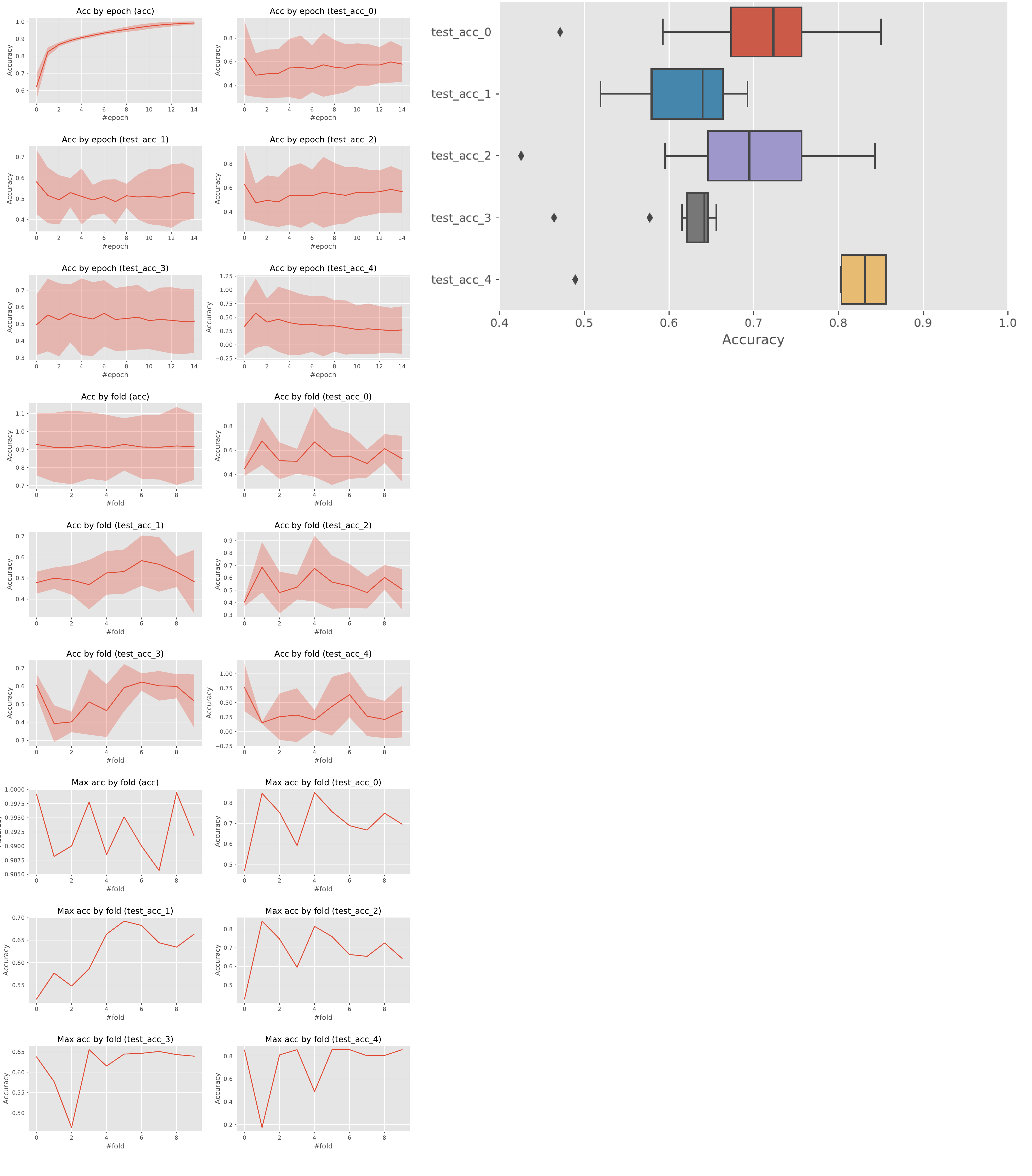}
\caption{max\_lr=0.005, lr\_decay=0.75}
\end{figure*}
\begin{figure*}[t]
	\centering
	\includegraphics[width=\linewidth]{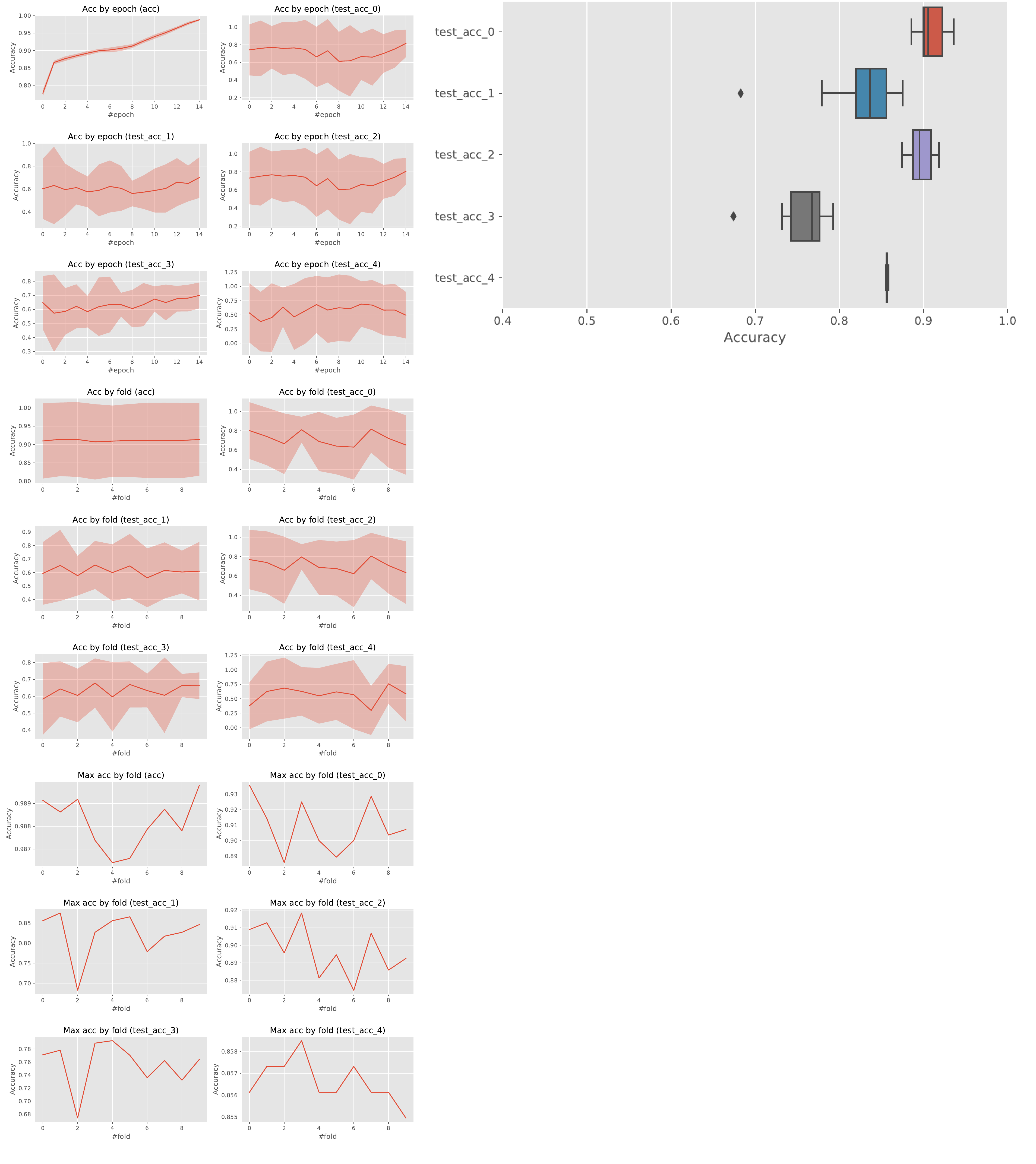}
	\caption{base\_lr=0.0001, max\_lr=0.005, cyclic\_mode=triangular2}
\end{figure*}
\begin{figure*}[t]
	\centering
	\includegraphics[width=\linewidth]{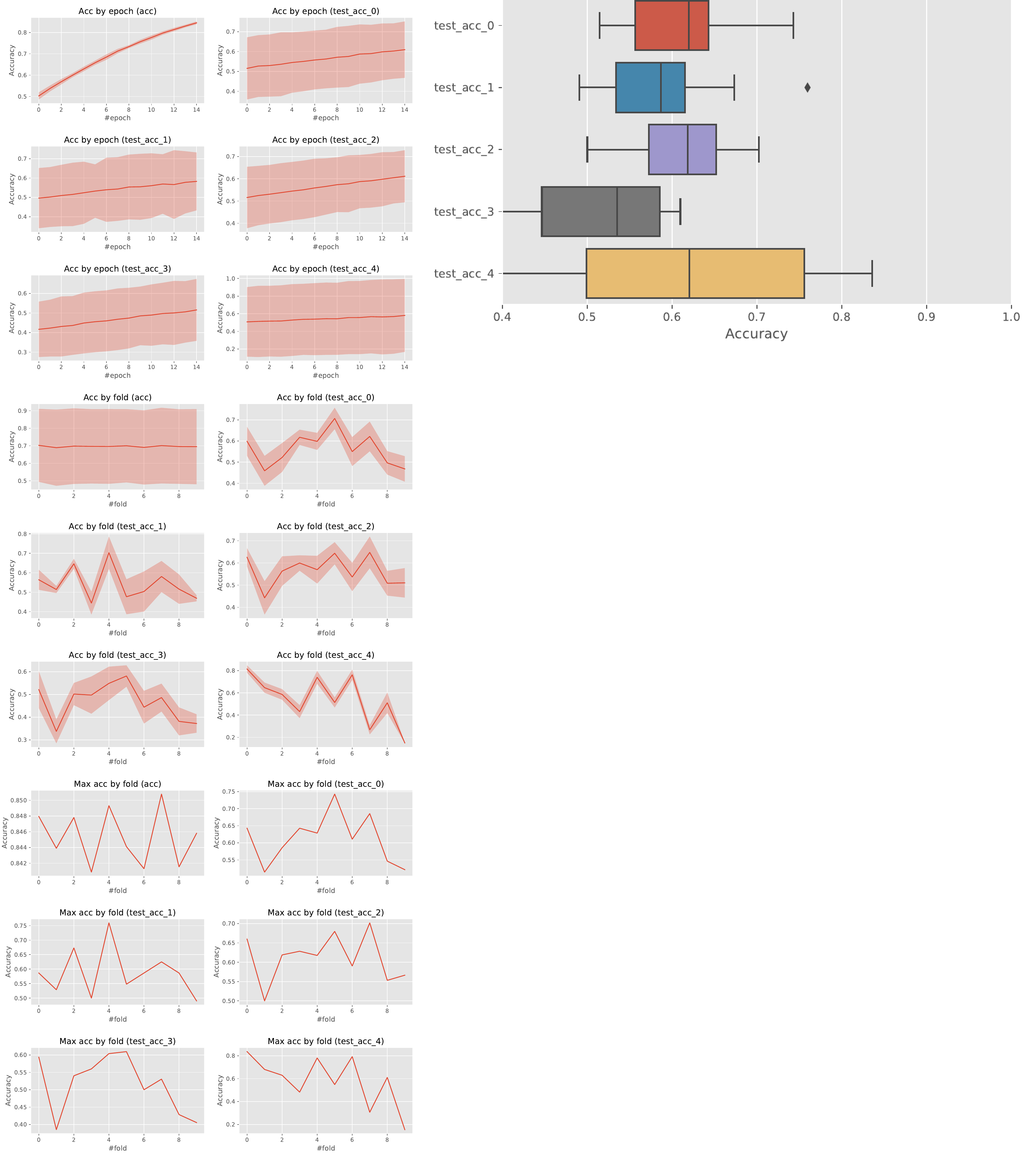}
	\caption{base\_lr=0.0005, max\_lr=0.001, cyclic\_mode=exp\_range}
\end{figure*}
\begin{figure*}[t]
	\centering
	\includegraphics[width=\linewidth]{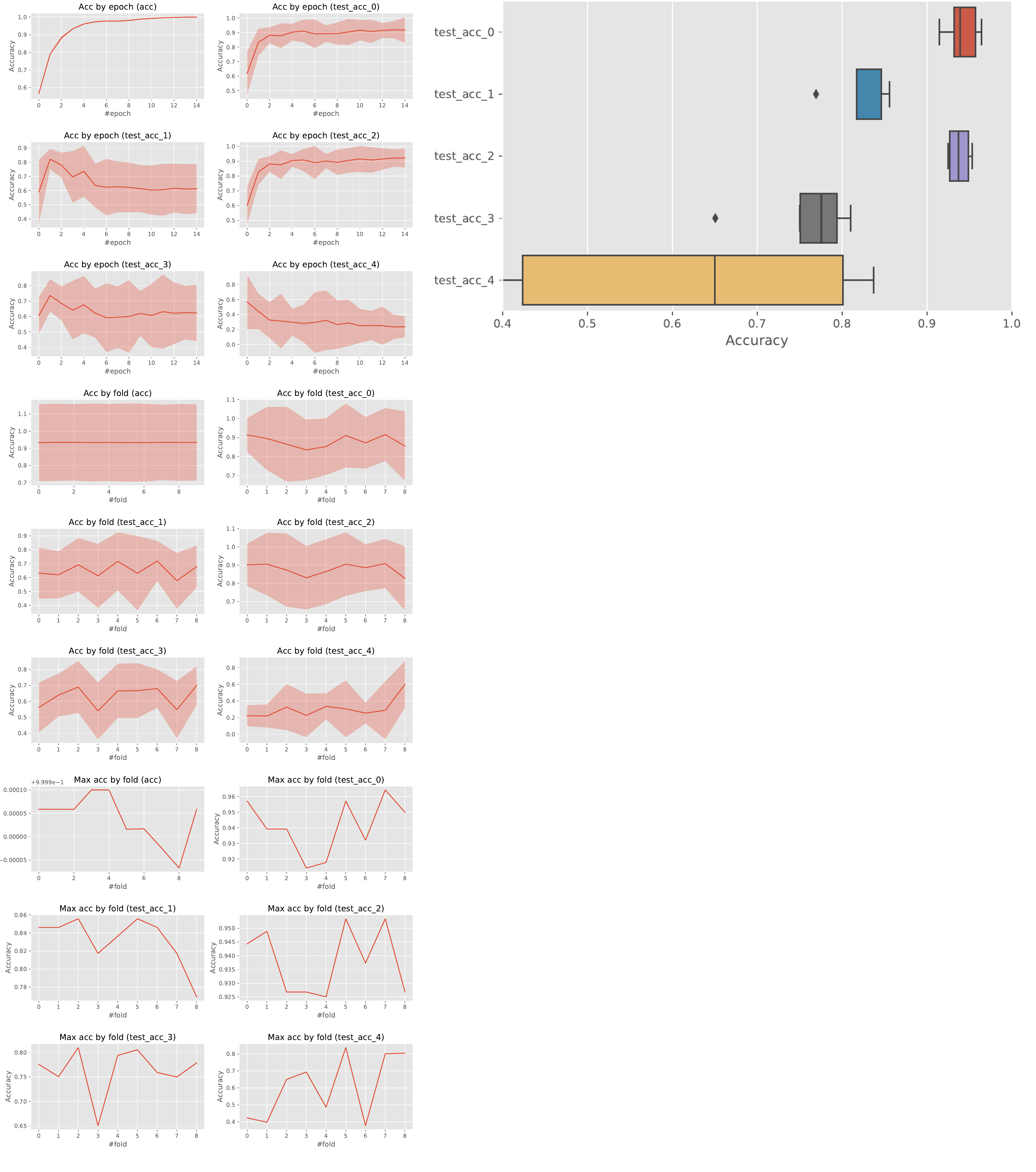}
	\caption{base\_lr=0.005, max\_lr=0.0001, cyclic\_mode=triangular}
\end{figure*}
\begin{figure*}[t]
	\centering
	\includegraphics[width=\linewidth]{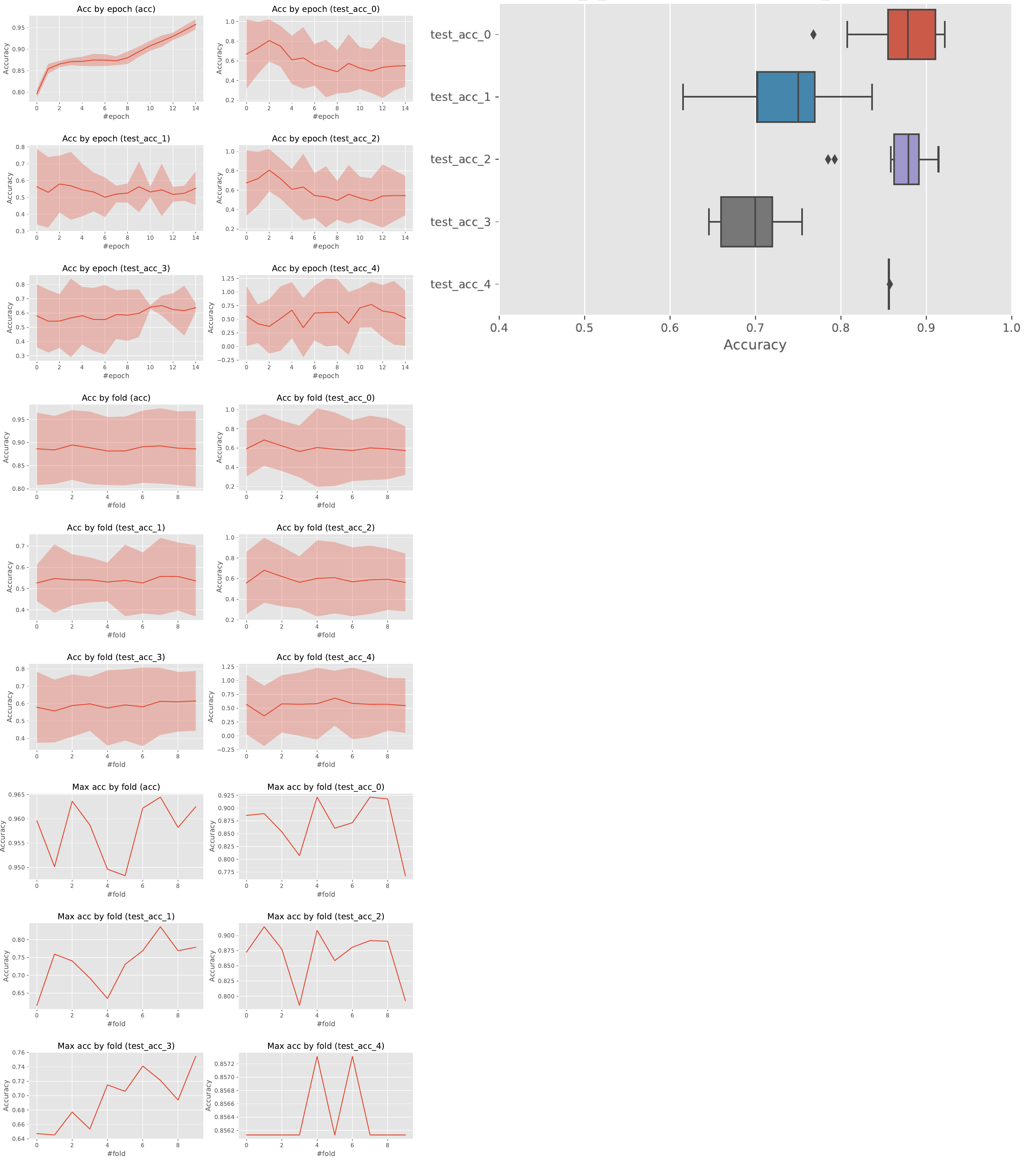}
	\caption{base\_lr=0.01, max\_lr=0.01, cyclic\_mode=triangular}
\end{figure*}
\end{document}